\documentclass[10pt,twocolumn,letterpaper]{article}

\usepackage{icb}
\usepackage{times}
\usepackage{epsfig}
\usepackage{graphicx}
\usepackage{amsmath}
\usepackage{amssymb}
\usepackage{xcolor,footnote,enumitem}
\usepackage{fancyhdr}
\usepackage[norule,symbol,perpage]{footmisc}
\usepackage{setspace}
\newcommand{\RNum}[1]{\uppercase\expandafter{\romannumeral #1\relax}}

\icbfinalcopy 

\ificbfinal\fi

\makeatletter
\def\ps@IEEEtitlepagestyle{
	\def\@oddfoot{\mycopyrightnotice}
	\def\@evenfoot{}
}
\def\mycopyrightnotice{
	{\hfill \footnotesize 978-1-7281-3640-0/19/\$31.00 \copyright 2019 IEEE\hfill}
}
\makeatother 

\begin{document}


\title{Polarimetric Thermal to Visible Face Verification\\ via Self-Attention Guided Synthesis}

\author{Xing Di$^1$,  Benjamin S. Riggan$^2$, Shuowen Hu$^2$, Nathaniel J. Short$^2$, Vishal M. Patel$^1$\\
$^1$Johns Hopkins University, 3400 N. Charles St, Baltimore, MD 21218, USA\\
$^2$U.S. Army CCDC Army Research Laboratory, 2800 Powder Mill Rd., Adelphi, MD 20783\\
{\tt\small xing.di@jhu.edu,\{benjamin.s.riggan.civ,shuowen.hu.civ\}@mail.mil}\\
{\tt\small short\_nathaniel@bah.com,vpatel36@jhu.edu}
}

\maketitle
\thispagestyle{empty}

\begin{abstract}
Polarimetric thermal to visible face verification entails matching two images that contain significant domain differences. Several recent approaches have attempted to synthesize visible faces from thermal images for cross-modal matching.  In this paper, we take a different approach in which rather than focusing only on synthesizing visible faces from thermal faces, we also propose to  synthesize thermal faces from visible faces.  Our intuition is based on the fact that thermal images also contain some discriminative information about the person for verification.  Deep features from a pre-trained Convolutional Neural Network (CNN) are extracted from the original as well as the synthesized images.  These features are then fused to generate a template which is then used for verification.  The proposed synthesis network is based on the self-attention generative adversarial network (SAGAN) which essentially allows efficient attention-guided image synthesis. Extensive experiments on the ARL polarimetric thermal face dataset demonstrate that the proposed method achieves state-of-the-art performance.

\end{abstract}
\let\thefootnote\relax\footnotetext{\mycopyrightnotice} 
\section{Introduction}

\begin{figure}
	\centering
	\includegraphics[width=1\linewidth]{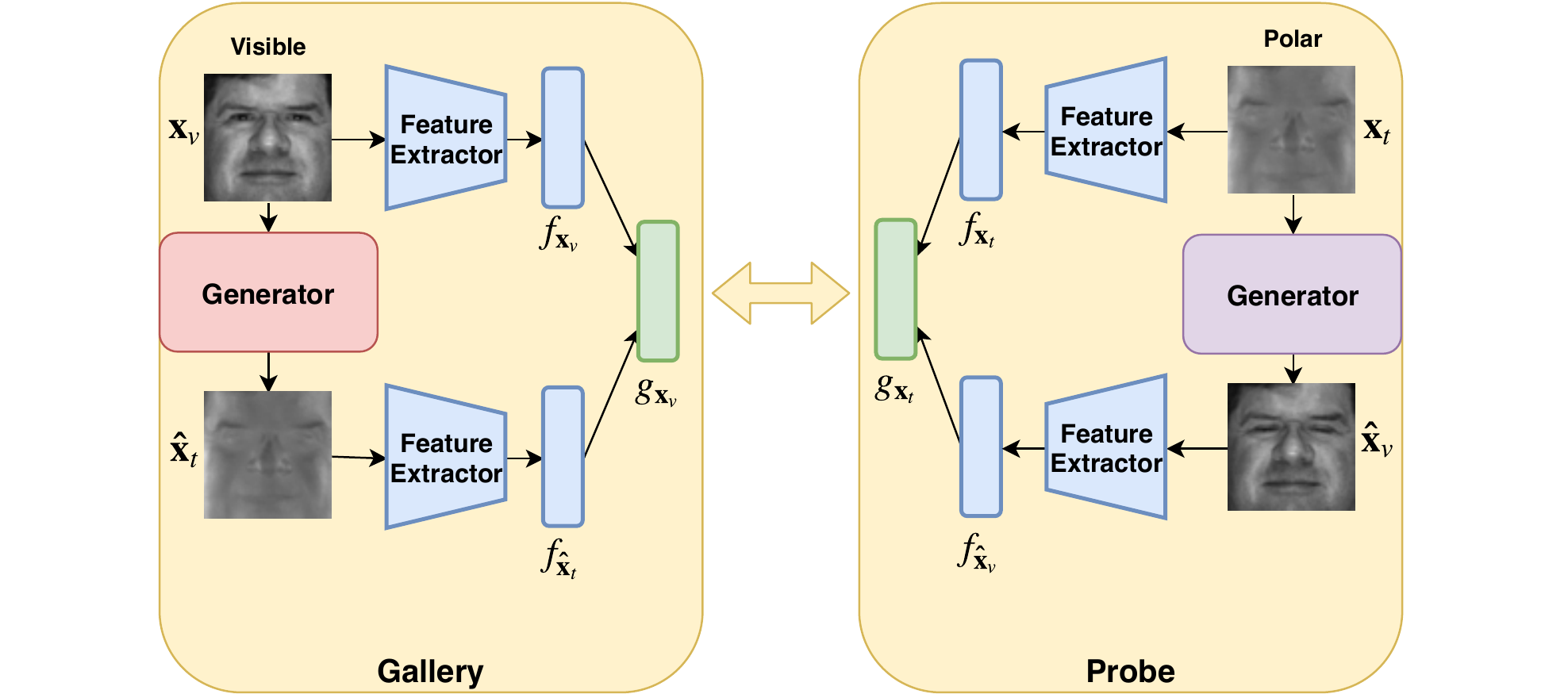}
	\caption{An overview of the proposed cross-modal face verification method.  Given a visible gallery image $\mathbf{x}_{v}$, a generator network is used to synthesize the corresponding thermal image $\hat{\mathbf{x}}_{t}$.  Similarly, given a polarimetric thermal probe image $\mathbf{x}_{t}$, a different generator network is used  to synthesize the corresponding visible image $\hat{\mathbf{x}}_{v}$.  Pre-trained CNNs are used to extract features from the original and the synthesized images.  These features are then fused to generate the gallery template $g_{\mathbf{x}_{v}}$ and the probe template $g_{\mathbf{x}_{t}}$.  Finally, the cosine similarity score between these feature templates is calculated for verification.}
	\label{fig:hfrgeneralour}
	\vspace{-5mm}
\end{figure}

Recognizing faces in low-light/night-time with that in normal (visible) conditions is a very difficult problem. Besides the challenges including expression, pose variance, etc.,  significant distribution change in different spectrum domain also causes the huge difficulty. Various thermal imaging modalities have been introduced in the literature to deal with this problem.  The  infrared spectrum can be divided into a reflection dominated region consisting of the near infrared (NIR) and shortwave infrared  (SWIR)  bands,  and  an  emission  dominated  thermal region consisting of the midwave infrared (MWIR) and longwave infrared (LWIR) bands \cite{riggan2016optimal}. It has been shown that polarimetric thermal imaging captures additional geometric and textural facial details compared to conventional thermal imaging \cite{hu2016polarimetric}.
Hence, the polarization-state information has been used to improve the performance of cross-spectrum face recognition \cite{hu2016polarimetric,riggan2016estimation,short2015improving,zhang2017generative,Riggan2018thermal,di2018polarimetric}.

 A  polarimetric, referred to as Stokes images, is composed of three channels:  S0, S1 and S2.  Here, S0 represents the conventional intensity only thermal image,  whereas S1 and S2 represent the horizontal/vertical and diagonal polarization-state information, respectively.   In polarimetric thermal to visible face verification, given a pair of visible and polarimetric thermal images, the goal is to determine whether these images correspond to the same person.  The large domain discrepancy between these images
makes the cross-spectrum matching problem very challenging.  Various methods have been proposed in the literature for cross-spectrum matching \cite{hu2016polarimetric,riggan2016estimation,short2015improving,zhang2017generative,Riggan2018thermal,klare2010heterogeneous,nicolo2012long, lezama2017not, bourlai2010cross,icme2018,DA_SPM}.  These approaches either attempt to synthesize visible faces from thermal faces or extract robust features from these modalities for cross-modal matching.

Hu \etal \cite{hu2015thermal} proposed a partial least squares regression (PLS) method for this cross-modal matching. Klare \etal \cite{klare2013heterogeneous} developed a generic framework based on the kernel prototype nonlinear similarity for cross-modal matching.  In \cite{thermalfacerecognition2012} PLS-based discriminant analysis approaches
 were used to correlate the thermal face signatures to the visible face signatures. 
 In addition, Riggan \etal \cite{riggan2016optimal} proposed a combination of PLS classifier with two different feature mapping approaches: Coupled Neural Network CpNN and Deep Perceptual Mapping (DPM) to utilize the features derived from the Stokes images for cross-modal face recognition.
 Recently, Iranmanesh \etal \cite{iranmanesh2018deep} proposed a two stream Deep Convolutional Neural Networks (DCNNs) (Vis-DCNN and Pol-DCNN) to learn a discriminative metric for this cross-domain verification.   
 Some of the other visible to thermal cross-modal matching methods include  \cite{Gurton:14,short2015exploiting}.

Various synthesis-based methods have also been proposed in the literature \cite{Riggan2018thermal,zhang2017generative,zhang2017tv,riggan2016estimation} to perform cross-modal mapping at the image level for direct use in existing visible-based matchers.  Riggan \etal \cite{riggan2016estimation} trained a regression network to estimate the mapping between features from both visible and thermal then reconstruct the visible face based on the estimated features. Zhang \etal \cite{zhang2017generative, zhang2018synthesis} leveraged  generative adversarial networks (GANs) to synthesize visible images from polarimetric thermal images.  Riggan \etal \cite{Riggan2018thermal} proposed a global and local region-based synthesis network to transform the thermal image into the visible spectrum. In a recent work \cite{di2018polarimetric} Di \etal  developed an attribute preserved adversarial network called AP-GAN to enhance the quality of the synthesized images. Zhang \etal \cite{zhang2018synthesis} introduced a multi-stream dense-residual encoder-decoder network, which implemented a feature-level fusion techniques to solve the problem.  

\begin{figure}[t]
	\centering
	\includegraphics[width=0.95\linewidth]{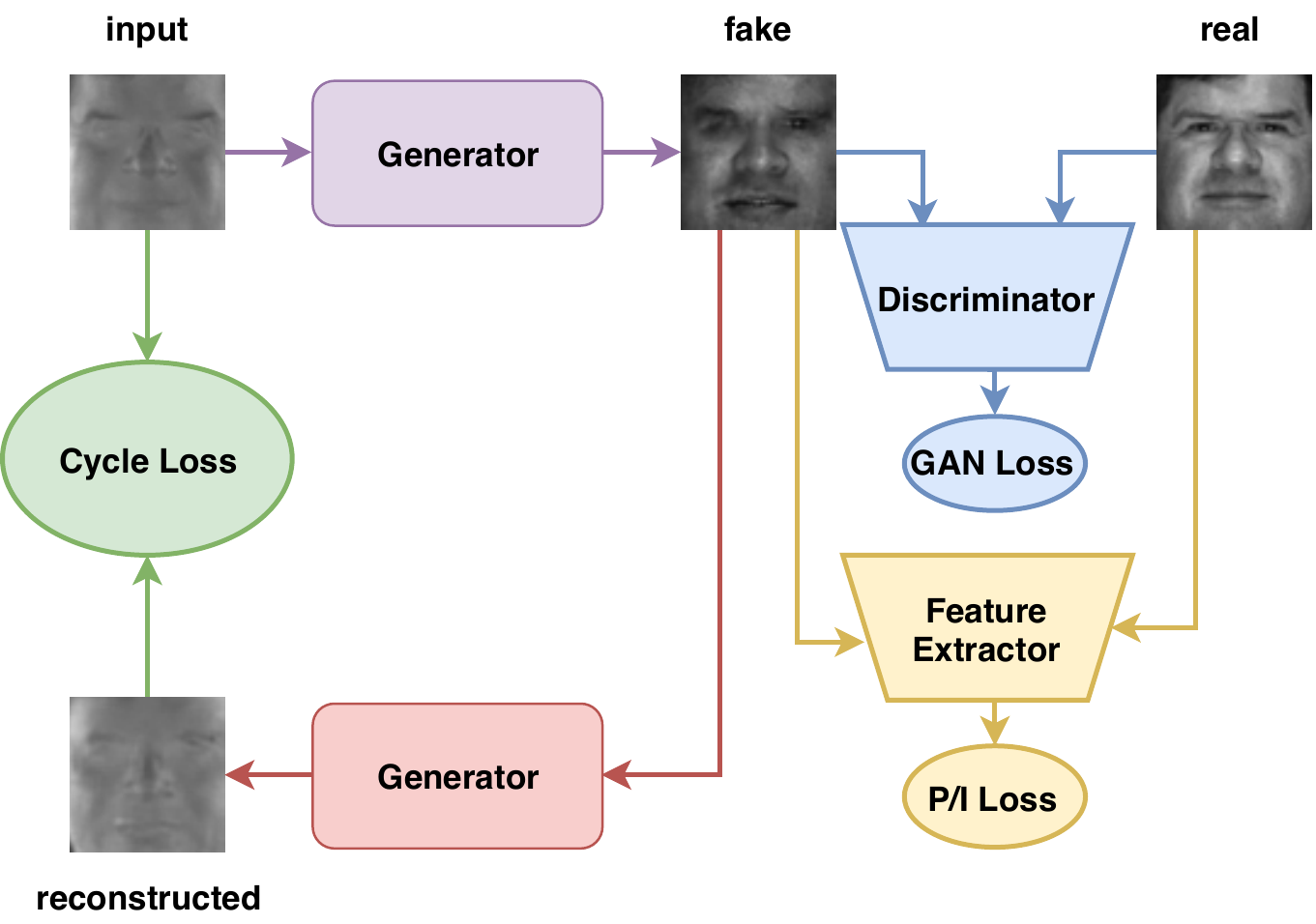}
	\caption{Self-attention guided synthesis of visible images from polarimetric thermal input. In order to minimize the domain gap between different modalities, the input thermal/visible images are directly mapped into the visible/thermal modality. In order to obtain the image level style, the pixel GAN loss (\textcolor{blue}{blue}) and cycle consistancy loss (\textcolor{green}{green}) are introduced. The feature-level semantic information is captured by the identity and  perceptual losses (\textcolor{yellow}{yellow}).  Similar architecture can also be used for synthesizing thermal images from visible images.}
	\label{fig:training}
	\vspace{-5mm}
\end{figure} 

In this paper, we take a different approach to the problem of thermal to visible matching by exploring the complementary information of different modalities. Figure~\ref{fig:hfrgeneralour} gives an overview of the proposed approach. Given a thermal-visible pair ($\mathbf{x}_{t}, \mathbf{x}_{v}$), these  images are first transformed into their spectrum counterparts using two trained generators as $\mathbf{\hat{x}}_{v} = G_{t\rightarrow v}(\mathbf{x}_{t})$, $\mathbf{\hat{x}}_{t} = G_{v\rightarrow t}(\mathbf{x}_{v})$. Then a feature extractor network $Feat$, in particular the VGG-Face model \cite{parkhi2015deep},  is used to extract features $f_{\mathbf{x}_{t}} = Feat(\mathbf{x}_{t})$, $f_{\mathbf{\hat{x}}_{v}} = Feat(\mathbf{\hat{x}}_{v})$, $f_{\mathbf{x}_{v}} = Feat(\mathbf{x}_{v})$, and $f_{\mathbf{\hat{x}}_{t}} = Feat(\mathbf{\hat{x}}_{t})$.  These features are then fused to generate the gallery template $g_{\mathbf{x}_{v}}= (f_{\mathbf{x}_{v}} + f_{\mathbf{\hat{x}}_{t}})/2$ and the probe template $g_{\mathbf{x}_{t}}=(f_{\mathbf{x}_{t}} + f_{\mathbf{\hat{x}}_{v}})/2$. Finally, the cosine similarity score between these feature templates is calculated for verification.

\begin{figure*}[!htb]
	\centering
	\includegraphics[width=0.95\linewidth]{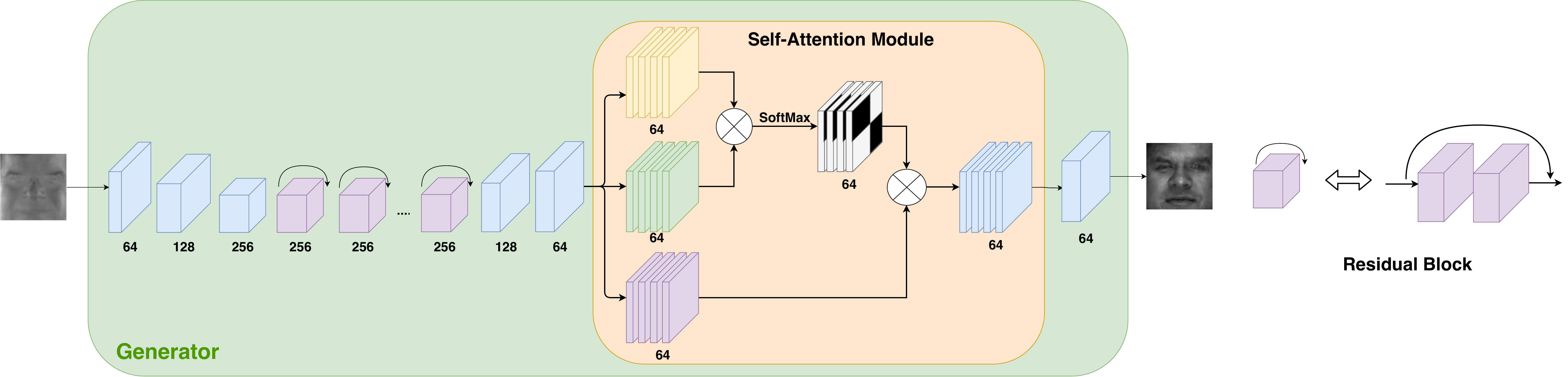}
	\caption{The proposed self-attention module-based generator architecture. }
	\label{fig:generator}
	\vspace{-5mm}
\end{figure*}

Note that CycleGAN-based networks \cite{CycleGAN2017} can be used to train these generators. However, experiments have shown that CycleGAN often fails to capture the geometric or struc- tural patterns around the eye and mouth regions.  One possible reason could be that the network relies heavily on convolutions to model the dependencies across different image regions. The long range dependencies are not well captured by the local receptive field of convolutional layers \cite{zhang2018self}. For improvement, we adopt the self-attention techniques \cite{cheng2016long,parikh2016decomposable,vaswani2017attention} from SAGAN \cite{zhang2018self}. The self-attention module is applied right before the last convolutional layer of the generator and the discriminator. Given the feature maps, this module learns the attention maps by itself with a softmax function and then the learned attention maps are multiplied with the feature maps to output the self-attention guided feature maps. In addition, the generator is optimized by an objective function consisting of the adversarial loss \cite{goodfellow2014generative},  $L_{1}$ loss, perceptual loss \cite{johnson2016perceptual},  identity loss \cite{zhang2017generative} and cycle-consistency loss \cite{CycleGAN2017}. The entire synthesis framework is shown in Figure~\ref{fig:training}.

To summarize, the following are our main contributions:
\begin{itemize}[noitemsep]
	
	\item  A novel cross-spectral face verification framework is proposed in which a self-attention guided GAN  is developed for synthesizing visible faces from the thermal and thermal faces from the visible.
	
	\item A novel self-attention module \cite{zhang2018self} based cycle-consistent \cite{CycleGAN2017} generator and pixel patch discriminator \cite{pix2pix2017} are proposed.
	
	\item 	Extensive experiments are conducted on the ARL Facial Database \cite{hu2016polarimetric} and  comparisons are  performed  against  several  recent  state-of-the-art approaches. Furthermore, an ablation study is conducted to demonstrate the effectiveness of the fusion approach proposed in this paper.  
	
\end{itemize}

\section{Proposed Method}

In this section, we discuss details of the proposed self-attention guided synthesis method. In particular, we discuss the proposed generator and the discriminator networks as well as the loss functions used to train the network. The overall framework is shown in Figure~\ref{fig:training}. Given an input image from one modality (thermal as shown), it is first synthesized into the other modality (i.e. visible) using the proposed self-attention module-based generator. Then another generator with similar architecture is used to synthesize it back from the visible domain to the original thermal domain. In order to achieve the reconstruction back to original modality,  these generators are trained using the cycle-consistency loss \cite{CycleGAN2017,yi2017dualgan}. In order to minimize the domain gap between the fake (i.e. synthesized) and real images, a patch-based pixel GAN loss is also introduced \cite{pix2pix2017}. Furthermore, the semantic and identity  information are captured by minimizing the perceptual and identity loss \cite{di2017gp}, respectively.

\subsection{Generator}
An encoder-decoder type of generator which is inspired by the residual network (He \etal \cite{he2016deep}) and SAGAN (Han \etal \cite{zhang2018self}) is adopted in this work.  In order to prevent the vanishing gradient problem, the residual block is implemented after a sequence of convolutional layers. For each residual block shown in Figure~\ref{fig:generator}, it consists of two convolutional layers followed by batch-normalization and relu layers. In order to involve the facial long-range dependency information, we adopt the self-attention module into the generator. Self-attention module was proposed by Han \etal in SAGAN \cite{zhang2018self} which allows attention-driven, long-range dependency modeling for general image generation tasks. In our work, the self-attention module is inserted right before the last convolutional layer of the generator. The self-attention module, shown as in Figure~\ref{fig:generator}, consists of two components: feature maps and attention maps. The feature maps are generated by a $1\times1$ convolutional layer working on the input features. The attention maps are generated by the elementwise multiplication of two $1\times1$ convolutional features followed by the softmax function. Finally, this module outputs the elementwise multiplication of feature maps and attention maps.

The self-attention module-based generator architecture is shown in Figure~\ref{fig:generator}.  This generator architecture is consists of the following components:

\noindent
CBR(64)-CBR(128)-CBR(256)-Res(256)-Res(256)-Res(256)-Res(256)-Res(256)-DBL(128)-DBL(64)-SA(64)-CT(3),

\noindent
where C stands for the convolutional layer (stride 2, kernel-size 4, and padding size 1). B and R stand for batch-normalization layer and relu layer, respectively.  Res is the residual block \cite{he2016deep}, D denotes  the deconvolutional layer (stride 2, kernel-size 4 and padding size 1),  L is the leaky relu layer, SA is the self-attention module, and T is the tanh function layer. The numbers inside the parenthesis denote  the number of channels corresponding to the output feature maps.

\subsection{Discriminator}

\begin{figure}[t]
	\centering
	\includegraphics[width=.95\linewidth]{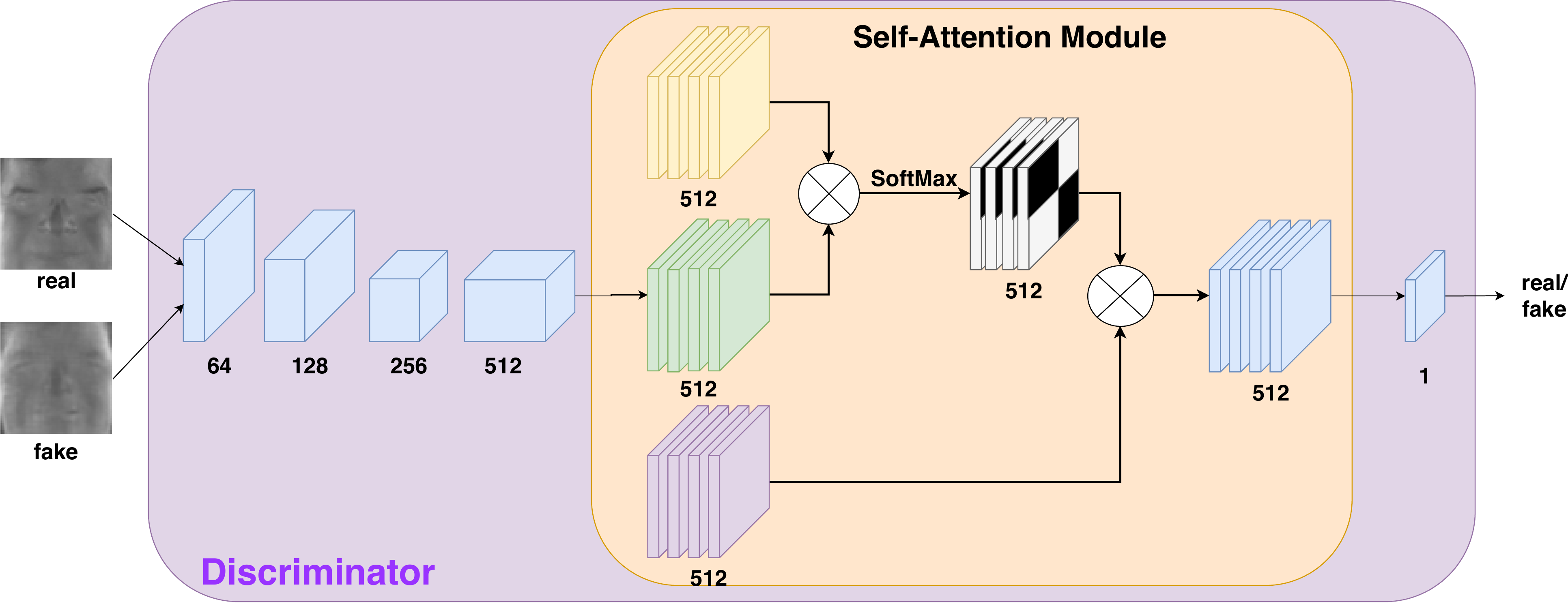}
	\caption{The architecture of the proposed discriminator.}
	\label{fig:discriminator}
	\vspace{-5mm}
\end{figure}

Motivated by pixel GAN \cite{pix2pix2017}, a patch-based discriminator is leveraged in the proposed method and is trained iteratively with the generator. In addition, in order to improve the stability of training, we adopt the spectral normalization to the discriminator  \cite{miyato2018spectral}. Compared to the other normalization techniques, spectral normalization does not require extra hyper-parameter tuning and has a relatively small the computational cost. Similarly, in order to capture the long-range dependency information, a self-attention module is added before the last convolutional layer in the discriminator. The discriminator, as shown in Figure~\ref{fig:discriminator}, consists of the following components:

\noindent
CLSn(64)-CLSn(128)-CLSn(256)-CLSn(512)-CLS(n512)-SA(512)-CS(1),

\noindent
where Sn stands for the spectral normalization layer. C, L, SA, S stand for the convolutional layer, leaky relu layer, self-attention module and sigmoid function, respectively. The numbers inside the parenthesis denote  the number of channels corresponding to the output feature maps.

\begin{figure*}[t]
	\centering
	\begin{minipage}{.45\textwidth}
		\centering
		\includegraphics[width=0.9\linewidth]{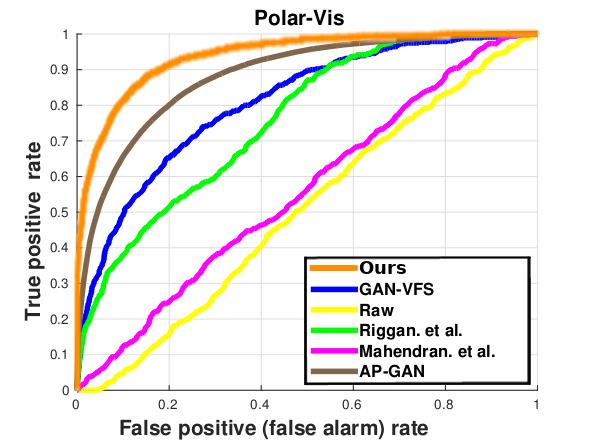}\\
		(a)
	\end{minipage}
	\begin{minipage}{.45\textwidth}
		\centering
		\includegraphics[width=0.9\linewidth]{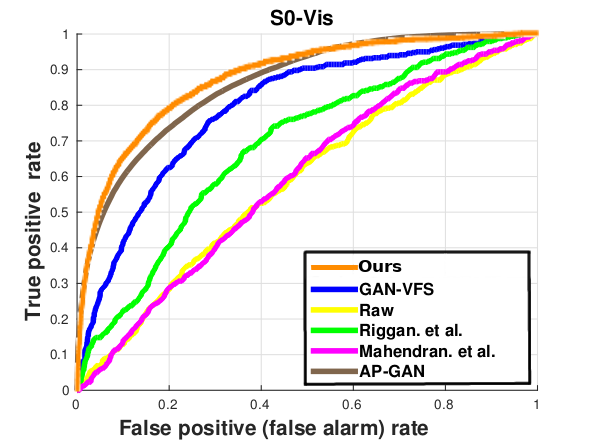}\\
		(b)
	\end{minipage}
	\caption{The ROC curve comparison on Protocol \RNum{1} with several state-of-the-art methods: GAN-VFS \cite{zhang2017generative}, Riggan \etal \cite{riggan2016estimation} Mahendran \etal \cite{mahendran2015understanding}, AP-GAN \cite{di2018polarimetric}. (a) The performance on Polar-Visible verification. (b) The performance  on S0-Visible verification.}
	\label{fig:Comparison_Figure}
	\vspace{-2mm}
\end{figure*}

\begin{figure*}
	\centering
	\includegraphics[width=.9\linewidth]{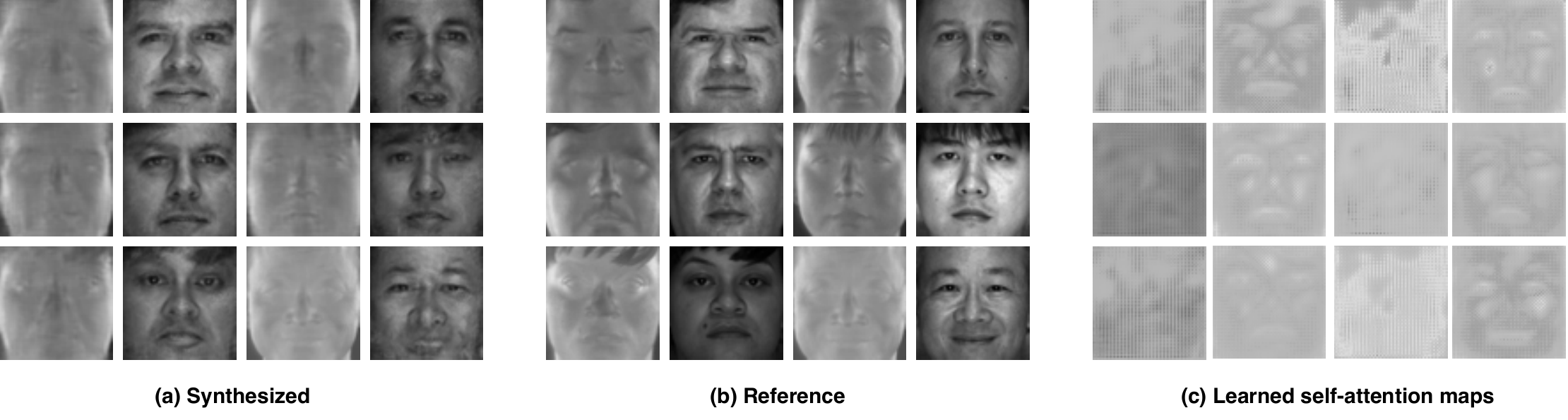} \\
	\caption{(a) Sample synthesized results on both visible and thermal modalities. (b) Reference images. (c) The learned self-attention feature maps. Images corresponding to different modality are shown in different columns.}
	\label{fig:compareprotocol1}
	\vspace{-5mm}
\end{figure*}

\subsection{Objective Function}
Given a set of thermal images $\mathbf{X}_{t}=\{x_{t}^{i}\}_{i=1}^{N}$ and another set of visible images $\mathbf{X}_{v}=\{x_{v}^{i}\}_{i=1}^{N}$, the generator and discriminator networks are optimized iteratively by minimizing the following loss functions
\begin{equation}\label{objective function}
\begin{split}
\nonumber \mathcal{L}_{} = \mathcal{L}_{GAN}(G_{t\rightarrow v}, D_{v}, \mathbf{X}_{t},\mathbf{X}_{v}) + \mathcal{L}_{GAN}(G_{v\rightarrow t}, D_{t}, \mathbf{X}_{v},\mathbf{X}_{t}) \\ 
+ \lambda_{P}\mathcal{L}_{P}(G_{t\rightarrow v}, \mathbf{X}_{t},\mathbf{X}_{v}) + \lambda_{P}\mathcal{L}_{P}(G_{v\rightarrow t}, \mathbf{X}_{v},\mathbf{X}_{t}) \\
+ \lambda_{I}\mathcal{L}_{I}(G_{t\rightarrow v}, \mathbf{X}_{t},\mathbf{X}_{v}) + \lambda_{I}\mathcal{L}_{I}(G_{v\rightarrow t}, \mathbf{X}_{v},\mathbf{X}_{t}) \\
+ \lambda_{1}\mathcal{L}_{1}(G_{t\rightarrow v}, \mathbf{X}_{t},\mathbf{X}_{v}) + \lambda_{1}\mathcal{L}_{1}(G_{v\rightarrow t}, \mathbf{X}_{v},\mathbf{X}_{t}) \\
+\mathcal{L}_{cycle}(G_{t\rightarrow v}, G_{v\rightarrow t}, \mathbf{X}_{t},\mathbf{X}_{v}),
\end{split}
\end{equation}
where $\mathcal{L}_{GAN}(G_{t\rightarrow v}, D_{v}, \mathbf{X}_{t},\mathbf{X}_{v}), \mathcal{L}_{GAN}(G_{v\rightarrow t}, D_{t}, \mathbf{X}_{v},\mathbf{X}_{t})$ are the adversarial losses for two generators - one for synthesizing visible from thermal ($G_{t\rightarrow v}$) and the other for synthesizing thermal from visible ($G_{v\rightarrow t}$). Similarly, $\mathcal{L}_{P}$ is the perceptual loss, $\mathcal{L}_{I}$ is the identity loss, $\mathcal{L}_{1}$ is the loss based on the L1-norm between the target and the synthesized image, and $\lambda_{P},\lambda_{I},\lambda_{1} $ are the weights for perceptual loss, identity loss and L1 loss, respectively.

\subsubsection{Adversarial Loss} 
Similar to the Cycle-GAN work \cite{CycleGAN2017}, there are two kinds adversarial losses. One $\mathcal{L}_{GAN}(G_{t\rightarrow v}, D_{v}, \mathbf{X}_{t},\mathbf{X}_{v})$ for synthesizing visible image from thermal image and the other $\mathcal{L}_{GAN}(G_{v\rightarrow t}, D_{t}, \mathbf{X}_{v},\mathbf{X}_{t})$ for synthesizing thermal image from visible image. Both are defined as follows:
\begin{equation}\label{adversarial loss}
\begin{split}
\mathcal{L}_{GAN}(G_{t\rightarrow v}, D_{v}, \mathbf{X}_{t},\mathbf{X}_{v}) = \mathbb{E}_{\mathbf{x}_{v}\sim \mathbf{X}_{v}} [\log D_{v} (\mathbf{x}_{v})],  \\ 
+ \mathbb{E}_{\mathbf{x}_{t}\sim \mathbf{X}_{t}}[\log (1-D_{v}(G_{t\rightarrow v}(\mathbf{x}_{t}))]\\
\mathcal{L}_{GAN}(G_{v\rightarrow t}, D_{t}, \mathbf{X}_{v},\mathbf{X}_{t}) = \mathbb{E}_{\mathbf{x}_{t}\sim \mathbf{X}_{t}} [\log D_{t} (\mathbf{x}_{t})]  \\ 
+ \mathbb{E}_{\mathbf{x}_{v}\sim \mathbf{X}_{v}}[\log (1-D_{t}(G_{v\rightarrow t}(\mathbf{x}_{v}))]
,\end{split}
\end{equation}
where $D_{v}$ and $D_{t}$ are discriminators for visible and thermal modality, respectively. In addition, $G_{v\rightarrow t}$ and $G_{v\rightarrow t}$ are two generators for synthesizing thermal image from visible and synthesizing visible image from thermal, respectively. 

\subsubsection{Cycle-Consistency Loss}
A cycle-consistency constraint is also imposed in our approach \cite{CycleGAN2017,yi2017dualgan} (see Figure~\ref{fig:training} green portion). Taking thermal to visible synthesis as an example, we introduce one mapping from thermal to visible $G_{t\rightarrow v}$ and train it according to the same GAN loss $\mathcal{L}_{GAN}(G_{t\rightarrow v}, D_{v}, \mathbf{X}_{t},\mathbf{X}_{v})$. We then require another mapping from  thermal to visible and back to thermal which reproduces the original
sample, thereby enforcing cycle-consistency. In other words, we want $G_{v\rightarrow t}(G_{t\rightarrow v}(\mathbf{x}_{t})) \sim \mathbf{x}_{t}$ and $G_{t\rightarrow v}(G_{v\rightarrow t}(\mathbf{x}_{v})) \sim \mathbf{x}_{v}$. This is done by imposing an $L_{1}$ penalty on the reconstruction error, which
is referred to as the cycle-consistency loss.  It is defined as follows:

\begin{equation}\label{Cycle-consistency loss}
\begin{split}
\mathcal{L}_{cycle}(G_{t\rightarrow v}, G_{v\rightarrow t}, \mathbf{X}_{t},\mathbf{X}_{v}) = \\ \mathbb{E}_{\mathbf{x}_{t}\sim \mathbf{X}_{t}}[\|G_{v\rightarrow t}(G_{t\rightarrow v}(\mathbf{x}_{t}))-\mathbf{x}_{t}\|_{1}] \\
+ \mathbb{E}_{\mathbf{x}_{v}\sim \mathbf{X}_{v}}[||G_{t\rightarrow v}(G_{v\rightarrow t}(\mathbf{x}_{v}))-\mathbf{x}_{v}\|_{1}].
\end{split}
\end{equation}

\subsubsection{Perceptual, Identity and L1 Loss Functions}
These loss functions can be implemented when we have supervised pairwise data $\{(\mathbf{x}^{i}_{t}, \mathbf{x}^{i}_{v})\}_{i=1}^{N}$, where $\mathbf{x}^{i}_{t} \in \mathbf{X}_{t} $ and $\mathbf{x}^{i}_{v} \in \mathbf{X_{v}}$,  during training.  The L1 loss is defined as below:
\begin{equation}\label{l1 loss}
\begin{split}
\mathcal{L}_{1}(G_{t\rightarrow v}, \mathbf{x}^{i}_{t},\mathbf{x}^{i}_{v})  = \|G_{t\rightarrow v}(\mathbf{x}^{i}_{t}) - \mathbf{x}^{i}_{v}\|_{1}\\
\mathcal{L}_{1}(G_{v\rightarrow t}, \mathbf{x}^{i}_{v},\mathbf{x}^{i}_{t}) = \|G_{v\rightarrow t}(\mathbf{x}^{i}_{v}) - \mathbf{x}^{i}_{t}\|_{1}.
\end{split}
\end{equation}
In order to minimize the perceptual and identity information \cite{johnson2016perceptual, di2017gp}, we implement the perceptual and identity loss functions as follows
\begin{equation}\label{perceptual loss}
\begin{split}
\nonumber \mathcal{L}_{p}(G_{t\rightarrow v}, F_{p}, \mathbf{x}^{i}_{t},\mathbf{x}^{i}_{v})  = [\|F_{p}(G_{t\rightarrow v}(\mathbf{x}^{i}_{t})) - F_{p}(\mathbf{x}^{i}_{v})\|_{1}]\\
\mathcal{L}_{p}(G_{v\rightarrow t}, F_{p}, \mathbf{x}^{i}_{v},\mathbf{x}^{i}_{t}) = [\|F_{p}(G_{v\rightarrow t}(\mathbf{x}^{i}_{v})) - F_{p}(\mathbf{x}^{i}_{t})\|_{1}]\\
\mathcal{L}_{I}(G_{t\rightarrow v}, F_{I}, \mathbf{x}^{i}_{t},\mathbf{x}^{i}_{v})  = [\|F_{I}(G_{t\rightarrow v}(\mathbf{x}^{i}_{t})) -  F_{I}(\mathbf{x}^{i}_{v})\|_{1}]\\
\mathcal{L}_{I}(G_{v\rightarrow t}, F_{I}, \mathbf{x}^{i}_{v},\mathbf{x}^{i}_{t}) = [\|F_{I}(G_{v\rightarrow t}(\mathbf{x}^{i}_{v})) -  F_{I}(\mathbf{x}^{i}_{t})\|_{1}],
\end{split}
\end{equation}
where $F_{I}$ and $F_{p}$ are two off-the-shelf pretrained networks for extracting features. Since deeper features in hierarchical deep networks capture more semantic information, the output features $conv2_2$ and $conv4_2$ from the VGGFace pretrained network are used in the perceptual and the identity losses, respectively.

Note that if we omit the perceptual, identity and L1 loss functions which require pairwise supervised data, then one can also implement the proposed framework in completely unsupervised fashion.  In other words, the proposed framework is also applicable to the case where the paired data are not available during training.  

\begin{table*}[htp!]
	\centering
	\caption{Protocol \RNum{1} Verification performance comparisons among the baseline methods and the proposed method for both polarimetric thermal (Polar) and conventional thermal (S0) cases.}
	\begin{tabular}{|c|c|c|c|c|}
		\hline 
		Method & AUC (Polar) & AUC(S0) & EER(Polar) & EER(S0) \\ 
		\hline
		Raw & 50.35\% & 58.64\% & 48.96\% & 43.96\% \\ 
		\hline 
		Mahendran \etal \cite{mahendran2015understanding}& 58.38\%  & 59.25\% & 44.56\% & 43.56\% \\ 
		\hline 
		Riggan \etal \cite{riggan2016estimation} & 75.83\% & 68.52\% & 33.20\% & 34.36\% \\ 
		\hline 
		GAN-VFS \cite{zhang2017generative} & 79.90\% & 79.30\% & 25.17\% & 27.34\% \\ 
		\hline 
		Riggan \etal  \cite{Riggan2018thermal} & 85.42\% & 82.49\% & 21.46\% & 26.25\% \\ 
		\hline 
		AP-GAN \cite{di2018polarimetric} & $88.93\% \pm 1.54\%$ & $84.16\% \pm1.54\%$ & $19.02\% \pm 1.69\%$ & $23.90\% \pm 1.52\%  $\\ 
		\hline 
		Multi-stream GAN  \cite{zhang2018synthesis} &$\mathbf{96.03}$\% &85.74\%&$\mathbf{11.78}$\%&23.18\%\\
		\hline
		Ours  &  $93.68\% \pm 0.97\% $ & $\mathbf{89.20\% \pm 1.56\%}$ & $13.46\% \pm 1.92\%$  & $\mathbf{18.77 \% \pm 1.36\% }$ \\ 
		\hline 
	\end{tabular}
	\label{tb:protocol 1}
	\vspace{-2mm}
\end{table*}

\begin{table*}[htp!]
	\centering
	\caption{Protocol \RNum{2} Verification performance comparisons among the baseline methods and the proposed method for both polarimetric thermal (Polar) and conventional thermal (S0) cases.}
	\begin{tabular}{|c|c|c|c|c|}
		\hline 
		Method & AUC (Polar) & AUC(S0) & EER(Polar) & EER(S0) \\ 
		\hline 
		Raw & 66.85\% & 63.66\% & 37.85\% & 40.93\% \\ 
		\hline
		CycleGAN \cite{CycleGAN2017}(unsupervised)  &  $76.09\% \pm 1.49\%$ &$ 74.17\% \pm 1.34\%$  & $32.28\% \pm 1.68\%$  & $33.04\% \pm 1.39\%$ \\ 
		\hline 		
		ours(unsupervised \footnote{ $L_{1}$, Identity and Perceptual losses are removed for training on unsupervised setting})  & $86.92\% \pm 1.42\% $ & $80.02\% \pm 1.16\%$ & $21.51 \% \pm 1.24\% $  & $28.09\% \pm 1.04\% $ \\
		\hline 
		Pix2Pix \cite{pix2pix2017} & $93.66\% \pm 1.07\%$ & $85.09\% \pm 1.48\%$ & $13.73\% \pm 1.38\%$ & $23.12\% \pm 1.14\%$  \\
		\hline
		Pix2PixBEGAN \cite{pix2pix2017,berthelot2017began} & $92.16\% \pm 1.09\%$ & $83.69\% \pm 1.28\%$& $15.38\% \pm 1.45\%$ & $26.22\% \pm 1.16\%$  \\
		\hline
		CycleGAN \cite{CycleGAN2017} (supervised)  &  $93.11\% \pm 1.02\% $ & $87.29\% \pm 1.13\% $  & $15.19\% \pm 1.02\%$  & $20.99\% \pm 1.19\%$  \\ 
		\hline 
		Multi-stream GAN  \cite{zhang2018synthesis} &$\mathbf{98.00}$\% &--&$\mathbf{7.99}$\%&--\\
		\hline
		Ours  &  $96.41\% \pm 1.02\% $ & $\mathbf{91.49\% \pm 2.25\%}$  & $10.02\% \pm 0.03\%$  & $\mathbf{15.45\% \pm 2.31\%}$ \\ 
		\hline 
	\end{tabular}
	\label{tb:protocol 2}
	\vspace{-2mm}
\end{table*}

\section{Experimental Results}
The proposed method is evaluated on the ARL Multimodal Face Database \cite{hu2016polarimetric} which consists of polarimetric  (i.e. Stokes image) and visible images from Volume \RNum{1} \cite{hu2016polarimetric} and \RNum{2} \cite{zhang2018synthesis}. The Volume \RNum{1} data consists of images corresponding to 60 subjects.  On the other hand, the Volume \RNum{2} data consists of images from 51 subjects (81 subjects in total).  Similar to \cite{riggan2016estimation,zhang2017generative,zhang2018synthesis}, we evaluate the proposed method based on two protocols. For Protocol \RNum{1}, images corresponding to Range 1 from 30 subjects are used for training. The remaining 30 subjects' data are used for evaluation. 
For Protocol \RNum{2}, all images from 81 subjects are used for running experiments. Specifically, all images from Volume \RNum{1} and 25 subjects' images from Volume \RNum{2} are used for training, the remaining 26 subjects' images from Volume \RNum{2} are used for evaluation.  We repeat this process 5 times and report the average results.

We evaluate the face verification performance of proposed method and compare it with several recent works \cite{zhang2017generative,Riggan2018thermal,pix2pix2017,CycleGAN2017, zhang2018synthesis}. Moreover, the performance is evaluated based on the FC-7 layer of the  pretrained VGG-Face model \cite{parkhi2015deep} using the receiver operating characteristic (ROC) curve, Area Under the Curve (AUC) and Equal Error Rate (EER) measures. To summarize, the proposed method is evaluated on the following four experiments:\\
\noindent a) Conventional thermal (S0) to Visible (Vis) on Protocol \RNum{1}. \\
\noindent b) Polarimetric thermal (Polar) to Visible (Vis) on Protocol \RNum{1}. \\
\noindent c) Conventional thermal (S0) to Visible (Vis) on Protocol \RNum{2}. \\
\noindent d) Polarimetric thermal (Polar) to Visible (Vis) on Protocol \RNum{2}.

\subsection{Implementation}
In addition to the standard preprocessing as discussed in \cite{hu2016polarimetric}, two more preprocessing steps are used in the proposed method. First, the faces in the visible domain are detected by MTCNN \cite{mtcnn}. Then, a standard central crop method is used to crop the registered faces. Since the MTCNN is implementable on the visible images only, we use the same detected rectangle coordination to crop the S0, S1, S2 images. After preprocessing, all the images are scaled to be $224\times224$ and are saved as 16-bit PNG files.

The entire network is trained in Pytorch on a single Nvidia Titan-X GPU. The L1, perceptual and identity loss parameters are chosen as $\lambda_{1}$=10, $\lambda_{p}$ = 2, $\lambda_{I}$ = 0.2 respectively by a grid search. The ADAM \cite{kingma2014adam} is implemented as the optimization algorithm with parameter  betas = (0.5,0.999) and batch size is chosen as 8. The total epochs are 200 for Protocol \RNum{1} and 100 for Protocol \RNum{2}. For the first half epochs, we fix the learning rate as $lr=0.0002$ and for the remaining epochs, the learning rate was decreased by 1/100 (Protocol \RNum{1}) and 1/50 (Protocol \RNum{2}) after each epoch.

Once the generators are trained, they could be implemented on the given probe and gallery images as shown in Figure~\ref{fig:hfrgeneralour}.

\subsection{Comparison with state-of-the-art Methods}
Regarding Protocol \RNum{1}, we evaluate and compare the performance of the proposed method with recent state-of-the-art methods \cite{zhang2017generative,mahendran2015understanding,riggan2016estimation,Riggan2018thermal,di2018polarimetric, zhang2018synthesis}. Figure \ref{fig:Comparison_Figure} shows the evaluation performance for two different experimental settings, S0 (representing conventional thermal) and Polar separately. As can be seen from Figure~\ref{fig:compareprotocol1}, compared with the other state-of-the-art methods, the proposed method performs better and comparably to \cite{zhang2018synthesis}.  In addition, it can be observed that the performance corresponding to the Polar modality is always better than the S0 modality, which demonstrates the advantage of using the polarimetric thermal images than the conventional thermal images. The quantitative comparisons are shown in  Table \ref{tb:protocol 1}, and also demonstrate the effectiveness of the proposed method. Furthermore, Figure~\ref{fig:compareprotocol1} shows some  synthesized images.  As can be seen from this figure, the facial attributes and the identity information is preserved well. 
Furthermore, from Figure~\ref{fig:compareprotocol1}(c) we see that the learned self-attention maps corresponding to both visible and thermal images are always located on the facial attributes regions such as mouth, eyes, and nose. As a result, the proposed self-attention guided GAN is able to capture meaningful information from both modalities for synthesis.

Table~\ref{tb:protocol 2} compares the performance of several state-of-the-art image synthesis on Protocol \RNum{2}.  These include multi-stream GAN \cite{zhang2018synthesis},
Pix2Pix \cite{pix2pix2017}, CycleGAN \cite{CycleGAN2017}, and Pix2Pix-BEGAN \cite{berthelot2017began}. Note that most prior works have not reported their results on  Protocol \RNum{2} as it is based on a new extended dataset that was only recently made publicly available. Similar to Protocol \RNum{1}, the experiments are evaluated on two different settings - S0 and Polar separately. As can be seen from Table~\ref{tb:protocol 2}, the proposed method performs comparably to the most recent state-of-the art image synthesize algorithms.  In this table, we also report the unsupervised performance of different methods.  As expected, the supervised results outperform the unsupervised results with a large margin.  Furthermore, the proposed method in unsupervised setting performs better than the other compared method.  This clearly shows the significance of using self-attention module in our framework.

Note that a GAN-based multi-stream fusion method recently proposed in \cite{zhang2018synthesis} is a supervised method that is specifically designed for the polarimetric data.  The generator network consists of a multi-stream feature-level
fusion encoder-decoder network.  As a result, the performance of \cite{zhang2018synthesis} is slightly better than our method on the polar modality.  On the other hand, our method outperforms \cite{zhang2018synthesis} by a large margin when only the S0 modality is used as the input.  Our method can be viewed as a generic heterogeneous face recognition method.  The performance of our method can be improved by using more sophisticated generators and feature extractors.

\begin{figure*}[htp!]
	\centering
	\begin{minipage}{.45\textwidth}
		\centering
		\includegraphics[width=0.9\linewidth]{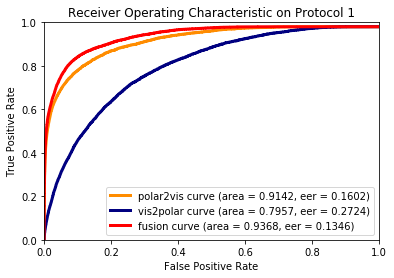}\\
	\end{minipage}
	\begin{minipage}{.45\textwidth}
		\centering
		\includegraphics[width=0.9\linewidth]{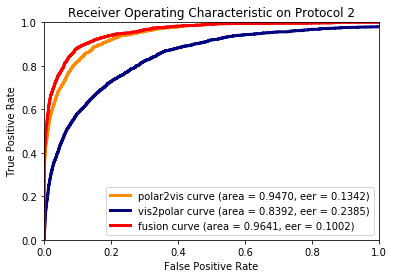}\\
	\end{minipage}
	\caption{The ROC curves corresponding to the proposed fusion method as well as individual modalities.}
	\label{fg: ablation study}
	\vspace{-5mm}
\end{figure*}

\subsection{Ablation Study Regarding Fusion}
In this section, we analyze the effectiveness of using fusion features in our method. In this ablation study, given polar (thermal) images $\mathbf{X}_{t}$ and visible images $\mathbf{X}_{v}$, we implement the following three experiments:

\noindent \textbf{polar2vis}: generate the visible images from the polar $\mathbf{\hat{X}}_{v} = G_{t\rightarrow v}(\mathbf{X}_{t})$, then verify based on the features from $(\mathbf{\hat{X}}_{v}, \mathbf{X}_{v})$.

\noindent \textbf{vis2polar}: generate the polar images from the visible $\mathbf{\hat{X}}_{t} = G_{v\rightarrow t}(\mathbf{X}_{v})$, then verify based on the feature from $(\mathbf{\hat{X}}_{t}, \mathbf{X}_{t})$.

\noindent \textbf{fusion}: generate the visible images from the polar $\mathbf{\hat{X}}_{v} = G_{t\rightarrow v}(\mathbf{X}_{t})$ and the polar images from the visible $\mathbf{\hat{X}}_{t} = G_{v\rightarrow t}(\mathbf{X}_{v})$, then verify the images based on the features from $((\mathbf{\hat{X}}_{t}+\mathbf{X}_{v})/2, (\mathbf{X}_{t} + \mathbf{\hat{X}}_{v})/2 )$.

This experiment will clearly show the significance of generating templates by fusing two features.  The ablation study is evaluated on both Protocol \RNum{1} and Protocol \RNum{2} and the results are shown in Figure~\ref{fg: ablation study}. Compared to the unimodal results, the fusion method significantly improves the performance on both protocols.  Also, the visible modality outperforms than the polar modality due to the reason that the  off-the-shelf VGGFace \cite{parkhi2015deep} feature extractor is pretrained on the visible face dataset.

\section{Conclusion}
We proposed a novel self-attention guided network for synthesizing thermal and visible faces for the task of cross-spectral face matching.  Given visible probe images, we synthesize the corresponding thermal images.   Similarly, given thermal probe images, we synthesize the visible images.  Features are then extracted from the original and the synthesized images.  Their fused feature representations are then used for verification.  The generators are based on the self-attention guided networks.  Various experiments on the ARL polarimetric thermal dataset were conducted to show the significance of the proposed approach.  Furthermore, an ablation study was conducted to show the improvements achieved by the proposed fusion approach.

Though we have only evaluated our approach on polarimetric thermal to visible face verification, in the future, we will evaluate the performance of this method on other heterogeneous face recognition tasks such as sketch to face matching \cite{PhotoSketch_ICPR2018}.

\section*{Acknowledgement}
This work was supported by the Defense Forensics \& Biometrics Agency (DFBA).  The authors would like to thank Mr. Bill Zimmerman and Ms. Michelle Giorgilli for their guidance and extensive discussions on this work.

{\small
\bibliographystyle{ieee}
\bibliography{icb2019}

\begin{thebibliography}{10}\itemsep=-1pt

\bibitem{berthelot2017began}
D.~Berthelot, T.~Schumm, and L.~Metz.
\newblock Began: boundary equilibrium generative adversarial networks.
\newblock {\em arXiv preprint arXiv:1703.10717}, 2017.

\bibitem{bourlai2010cross}
T.~Bourlai, N.~Kalka, A.~Ross, B.~Cukic, and L.~Hornak.
\newblock Cross-spectral face verification in the short wave infrared (swir)
  band.
\newblock In {\em Pattern Recognition (ICPR), 2010 20th International
  Conference on}, pages 1343--1347. IEEE, 2010.

\bibitem{cheng2016long}
J.~Cheng, L.~Dong, and M.~Lapata.
\newblock Long short-term memory-networks for machine reading.
\newblock {\em arXiv preprint arXiv:1601.06733}, 2016.

\bibitem{di2017gp}
X.~Di, V.~A. Sindagi, and V.~M. Patel.
\newblock Gp-gan: Gender preserving gan for synthesizing faces from landmarks.
\newblock In {\em 2018 24th International Conference on Pattern Recognition
  (ICPR)}, pages 1079--1084, Aug 2018.

\bibitem{di2018polarimetric}
X.~Di, H.~Zhang, and V.~M. Patel.
\newblock Polarimetric thermal to visible face verification via attribute
  preserved synthesis.
\newblock {\em IEEE International Conference on Biometrics: Theory,
  Applications, and Systems (BTAS)}, Oct. 2018.

\bibitem{goodfellow2014generative}
I.~Goodfellow, J.~Pouget-Abadie, M.~Mirza, B.~Xu, D.~Warde-Farley, S.~Ozair,
  A.~Courville, and Y.~Bengio.
\newblock Generative adversarial nets.
\newblock In {\em Advances in neural information processing systems}, pages
  2672--2680, 2014.

\bibitem{Gurton:14}
K.~P. Gurton, A.~J. Yuffa, and G.~W. Videen.
\newblock Enhanced facial recognition for thermal imagery using polarimetric
  imaging.
\newblock {\em Opt. Lett.}, 39(13):3857--3859, Jul 2014.

\bibitem{he2016deep}
K.~He, X.~Zhang, S.~Ren, and J.~Sun.
\newblock Deep residual learning for image recognition.
\newblock In {\em Proceedings of the IEEE conference on computer vision and
  pattern recognition}, pages 770--778, 2016.

\bibitem{hu2015thermal}
S.~Hu, J.~Choi, A.~L. Chan, and W.~R. Schwartz.
\newblock Thermal-to-visible face recognition using partial least squares.
\newblock {\em JOSA A}, 32(3):431--442, 2015.

\bibitem{hu2016polarimetric}
S.~Hu, N.~J. Short, B.~S. Riggan, C.~Gordon, K.~P. Gurton, M.~Thielke,
  P.~Gurram, and A.~L. Chan.
\newblock A polarimetric thermal database for face recognition research.
\newblock In {\em Proceedings of the IEEE Conference on Computer Vision and
  Pattern Recognition Workshops}, pages 119--126, 2016.

\bibitem{iranmanesh2018deep}
S.~M. Iranmanesh, A.~Dabouei, H.~Kazemi, and N.~M. Nasrabadi.
\newblock Deep cross polarimetric thermal-to-visible face recognition.
\newblock In {\em 2018 International Conference on Biometrics (ICB)}, pages
  166--173, Feb 2018.

\bibitem{pix2pix2017}
P.~Isola, J.-Y. Zhu, T.~Zhou, and A.~A. Efros.
\newblock Image-to-image translation with conditional adversarial networks.
\newblock {\em CVPR}, 2017.

\bibitem{johnson2016perceptual}
J.~Johnson, A.~Alahi, and L.~Fei-Fei.
\newblock Perceptual losses for real-time style transfer and super-resolution.
\newblock In {\em European Conference on Computer Vision}, pages 694--711.
  Springer, 2016.

\bibitem{thermalfacerecognition2012}
S.~S. Y. L. S.~D. Jonghyun~Choi, Shuowen~Hu.
\newblock Thermal to visible face recognition.
\newblock In {\em Proc.SPIE}, pages 8371 -- 8371 -- 10, 2012.

\bibitem{kingma2014adam}
D.~P. Kingma and J.~Ba.
\newblock Adam: A method for stochastic optimization.
\newblock In {\em International Conference on Learning Representations (ICLR)},
  2014.

\bibitem{klare2010heterogeneous}
B.~Klare and A.~K. Jain.
\newblock Heterogeneous face recognition: Matching nir to visible light images.
\newblock In {\em Pattern Recognition (ICPR), 2010 20th International
  Conference on}, pages 1513--1516. IEEE, 2010.

\bibitem{klare2013heterogeneous}
B.~F. Klare and A.~K. Jain.
\newblock Heterogeneous face recognition using kernel prototype similarities.
\newblock {\em IEEE transactions on pattern analysis and machine intelligence},
  35(6):1410--1422, 2013.

\bibitem{lezama2017not}
J.~Lezama, Q.~Qiu, and G.~Sapiro.
\newblock Not afraid of the dark: Nir-vis face recognition via cross-spectral
  hallucination and low-rank embedding.
\newblock In {\em 2017 IEEE Conference on Computer Vision and Pattern
  Recognition (CVPR)}, pages 6807--6816. IEEE, 2017.

\bibitem{mahendran2015understanding}
A.~Mahendran and A.~Vedaldi.
\newblock Understanding deep image representations by inverting them.
\newblock In {\em IEEE Conference on Computer Vision and Pattern Recognition},
  2015.

\bibitem{miyato2018spectral}
T.~Miyato, T.~Kataoka, M.~Koyama, and Y.~Yoshida.
\newblock Spectral normalization for generative adversarial networks.
\newblock In {\em International Conference on Learning Representations}, 2018.

\bibitem{nicolo2012long}
F.~Nicolo and N.~A. Schmid.
\newblock Long range cross-spectral face recognition: matching swir against
  visible light images.
\newblock {\em IEEE Transactions on Information Forensics and Security},
  7(6):1717--1726, 2012.

\bibitem{parikh2016decomposable}
A.~P. Parikh, O.~T{\"a}ckstr{\"o}m, D.~Das, and J.~Uszkoreit.
\newblock A decomposable attention model for natural language inference.
\newblock In {\em The Conference on Empirical Methods in Natural Language
  Processing}, 2016.

\bibitem{parkhi2015deep}
O.~M. Parkhi, A.~Vedaldi, A.~Zisserman, et~al.
\newblock Deep face recognition.
\newblock In {\em British Machine Vision Conference}, 2015.

\bibitem{DA_SPM}
V.~M. {Patel}, R.~{Gopalan}, R.~{Li}, and R.~{Chellappa}.
\newblock Visual domain adaptation: A survey of recent advances.
\newblock {\em IEEE Signal Processing Magazine}, 32(3):53--69, May 2015.

\bibitem{riggan2016optimal}
B.~S. Riggan, N.~J. Short, and S.~Hu.
\newblock Optimal feature learning and discriminative framework for
  polarimetric thermal to visible face recognition.
\newblock In {\em 2016 IEEE Winter Conference on Applications of Computer
  Vision (WACV)}, pages 1--7. IEEE, 2016.

\bibitem{Riggan2018thermal}
B.~S. Riggan, N.~J. Short, and S.~Hu.
\newblock Thermal to visible synthesis of face images using multiple regions.
\newblock In {\em IEEE Winter Conference on Applications of Computer Vision
  (WACV)}, 2018.

\bibitem{riggan2016estimation}
B.~S. Riggan, N.~J. Short, S.~Hu, and H.~Kwon.
\newblock Estimation of visible spectrum faces from polarimetric thermal faces.
\newblock In {\em Biometrics Theory, Applications and Systems (BTAS), 2016 IEEE
  8th International Conference on}, pages 1--7. IEEE, 2016.

\bibitem{icme2018}
B.~S. Riggan, N.~J. Short, M.~S. Sarfraz, S.~Hu, H.~Zhang, V.~M. Patel,
  S.~Rasnayaka, J.~Li, T.~Sim, S.~M. Iranmanesh, and N.~M. Nasrabadi.
\newblock Icme grand challenge results on heterogeneous face recognition:
  Polarimetric thermal-to-visible matching.
\newblock In {\em 2018 IEEE International Conference on Multimedia Expo
  Workshops (ICMEW)}, pages 1--4, July 2018.

\bibitem{short2015exploiting}
N.~Short, S.~Hu, P.~Gurram, and K.~Gurton.
\newblock Exploiting polarization-state information for cross-spectrum face
  recognition.
\newblock In {\em Biometrics Theory, Applications and Systems (BTAS), 2015 IEEE
  7th International Conference on}, pages 1--6. IEEE, 2015.

\bibitem{short2015improving}
N.~Short, S.~Hu, P.~Gurram, K.~Gurton, and A.~Chan.
\newblock Improving cross-modal face recognition using polarimetric imaging.
\newblock {\em Optics letters}, 40(6):882--885, 2015.

\bibitem{vaswani2017attention}
A.~Vaswani, N.~Shazeer, N.~Parmar, J.~Uszkoreit, L.~Jones, A.~N. Gomez,
  {\L}.~Kaiser, and I.~Polosukhin.
\newblock Attention is all you need.
\newblock In {\em Advances in Neural Information Processing Systems}, pages
  5998--6008, 2017.

\bibitem{PhotoSketch_ICPR2018}
L.~{Wang}, V.~{Sindagi}, and V.~{Patel}.
\newblock High-quality facial photo-sketch synthesis using multi-adversarial
  networks.
\newblock In {\em 2018 13th IEEE International Conference on Automatic Face
  Gesture Recognition (FG 2018)}, pages 83--90, May 2018.

\bibitem{yi2017dualgan}
Z.~Yi, P.~Tan, and M.~Gong.
\newblock Dualgan: Unsupervised dual learning for image-to-image translation.
\newblock In {\em ICCV}, 2017.

\bibitem{zhang2018self}
H.~Zhang, I.~Goodfellow, D.~Metaxas, and A.~Odena.
\newblock Self-attention generative adversarial networks.
\newblock {\em arXiv preprint arXiv:1805.08318}, 2018.

\bibitem{zhang2017generative}
H.~{Zhang}, V.~M. {Patel}, B.~S. {Riggan}, and S.~{Hu}.
\newblock Generative adversarial network-based synthesis of visible faces from
  polarimetrie thermal faces.
\newblock In {\em 2017 IEEE International Joint Conference on Biometrics
  (IJCB)}, pages 100--107, Oct 2017.

\bibitem{zhang2018synthesis}
H.~Zhang, B.~S. Riggan, S.~Hu, N.~J. Short, and V.~M. Patel.
\newblock Synthesis of high-quality visible faces from polarimetric thermal
  faces using generative adversarial networks.
\newblock {\em International Journal of Computer Vision: Special Issue on Deep
  Learning for Face Analysis}, 2019.

\bibitem{mtcnn}
K.~Zhang, Z.~Zhang, Z.~Li, and Y.~Qiao.
\newblock Joint face detection and alignment using multitask cascaded
  convolutional networks.
\newblock {\em IEEE Signal Processing Letters}, 23(10):1499--1503, Oct 2016.

\bibitem{zhang2017tv}
T.~Zhang, A.~Wiliem, S.~Yang, and B.~Lovell.
\newblock Tv-gan: Generative adversarial network based thermal to visible face
  recognition.
\newblock In {\em 2018 International Conference on Biometrics (ICB)}, pages
  174--181. IEEE, 2018.

\bibitem{CycleGAN2017}
J.-Y. Zhu, T.~Park, P.~Isola, and A.~A. Efros.
\newblock Unpaired image-to-image translation using cycle-consistent
  adversarial networkss.
\newblock In {\em Computer Vision (ICCV), 2017 IEEE International Conference
  on}, 2017.

\end{thebibliography}
}

\end{document}